\documentclass[sn-mathphys,Numbered]{sn-jnl}

\usepackage{graphicx}%
\usepackage{multirow}%
\usepackage{amsmath,amssymb,amsfonts}%
\usepackage{amsthm}%
\usepackage{mathrsfs}%
\usepackage[title]{appendix}%
\usepackage{xcolor}%
\usepackage{textcomp}%
\usepackage{manyfoot}%
\usepackage{booktabs}%
\usepackage{algorithm}%
\usepackage{algorithmicx}%
\usepackage{algpseudocode}%
\usepackage{listings}%
\usepackage{changepage} 
\usepackage{gensymb}
\usepackage{float}

\begin{document}
	
	\title[Article Title]{
        A Bio-Inspired Research Paradigm of Collision Perception Neurons Enabling Neuro-Robotic Integration: The LGMD Case
        }
	
	
	\author[1,2]{\fnm{Ziyan} \sur{Qin}}\email{ziyan9603@e.gzhu.edu.cn}
	\author[1]{\fnm{Jigen} \sur{Peng}}\email{jgpeng@gzhu.edu.cn}
	\author[2]{\fnm{Shigang} \sur{Yue}}\email{sy237@le.ac.uk}
    \author*[1]{\fnm{Qinbing} \sur{Fu}}\email{qifu@gzhu.edu.cn}
	
	\affil*[1]{\orgdiv{Machine Life and Intelligence Research Centre, School of Mathematics and Information Science},      \orgname{Guangzhou University}, \orgaddress{\street{No.230 Waihuan West Road},
			\city{Guangzhou}, \postcode{510006}, \country{China}}}
	
	\affil[2]{\orgdiv{School of Computing and Mathematical Sciences}, \orgname{University of Leicester}, \orgaddress{\street{University Road},
			\city{Leicester}, \postcode{LE1 7RH}, \country{UK}}}
	
	

	\maketitle
	
	\begin{adjustwidth}{1cm}{1cm} 
        Compared to human vision, locust visual systems excel at rapid and precise collision detection, despite relying on only hundreds of thousands of neurons organized through a few neuropils. 
        This efficiency makes them an attractive model system for developing artificial collision-detecting systems. 
        Specifically, researchers have identified collision-selective neurons in the locust's optic lobe, called lobula giant movement detectors (LGMDs), which respond specifically to approaching objects.
		
        Research upon LGMD neurons began in the early 1970s. Initially, due to their large size, these neurons were identified as motion detectors, but their role as looming detectors was recognized over time. 
        Since then, progress in neuroscience, computational modeling of LGMD's visual neural circuits, and LGMD-based robotics have advanced in tandem, each field supporting and driving the others. 
        Today, with a deeper understanding of LGMD neurons, LGMD-based models have significantly improved collision-free navigation in mobile robots including ground and aerial robots.
	
        This review highlights recent developments in LGMD research from the perspectives of neuroscience, computational modeling, and robotics. 
        It emphasizes a biologically plausible research paradigm, where insights from neuroscience inform real-world applications, which would in turn validate and advance neuroscience. 
        With strong support from extensive research and growing application demand, this paradigm has reached a mature stage and demonstrates versatility across different areas of neuroscience research, thereby enhancing our understanding of the interconnections between neuroscience, computational modeling, and robotics. 
        Furthermore, this paradigm would shed light upon the modeling and robotic research into other motion-sensitive neurons or neural circuits.
	\end{adjustwidth}
	
	\vspace{1cm}

\section*{Nomenclature}
    \begin{tabular}{ll}
	LGMD			&lobula giant movement detector\\
    DCMD            &descending contralateral movement detector\\
    TmA             &trans-medullary afferent neuron\\
    FFI             &feed-forward inhibition\\
    LMC             &lamina monopolar cell\\
    SIZ             &spike initiation zone\\
    DUB             &dorsal uncrossed bundle\\
    RF				&receptive field\\
    \end{tabular}

\section{Introduction}
    Visually guided collision avoidance is a common behavior in many sighted animal species. 
    Unlike humans, locusts can accurately perceive impending collisions using a neural circuit comprising only hundreds of thousands of neurons, interconnected through a few neuropils. 
    In adult locusts, at least 150,000 neurons contribute to collision detection—a significantly smaller number than invertebrates. 
    This compact neural architecture makes the locust visual system an ideal model for developing efficient and simplified collision detection mechanisms.
    
    Among insects, locusts excel at this ability, demonstrating the remarkable capacity to travel in swarms over long distances without collisions. 
    A bilateral pair of lobula giant movement detectors (LGMDs) in the locust's visual brain plays a key role in looming\footnote{visual movement induced by object approaching the eyes of an animal} perception. 
    Two specific LGMDs, LGMD1 and LGMD2, have been identified as responsible for this function. 
    They respond most strongly to approaching objects while distinguishing between different motion types, such as receding, translational, and whole-field shifting movements \cite{Schlotterer1977,Rind1997}.

    Anatomical research on the LGMDs and circuitry was inaugurated in the 1970s \cite{Oshea1976a,Oshea1976b,Oshea1976c,Oshea1976d}, while two distinct types of computational models of LGMD1 emerged in the 1990s \cite{Gabbiani1995,Rind1996a}. 
    The first type of computational models incorporates physical attributes, i.e., image size (angular size) and angular speed on the visual receptive field (RF), to represent the LGMD's response to looming objects, focusing on dendritic computations within the neuron.
    In contrast, the second type of models utilizes critical image cues and lateral inhibition within the signal processing pathway to LGMD1 to simulate its neural responses, enabling the distinction between approaching, receding, and translating stimuli.
    Although LGMD2 shows the similar looming selectivity\footnote{neuron responding most strongly to approaching rather than other kinds of visual movements} of LGMD1, computational modeling of LGMD2 has emerged late in recent years due to very limited biological findings, compared to LGMD1 \cite{Rind1997b}.
    
    In real-world scenarios, the angular size and speed of an approaching object on the RF are often unknown, unable to measure, which poses significant challenges for implementing the first type of computational model in complex robot scenarios interacting with dynamic environments. 
    Moreover, this model could only predict the LGMD1 response to approaching stimuli, without accounting for other types of motion like receding and translating. 
    In contrast, the second type of model has been successfully extended and demonstrated in real-world environments using a mobile robot system \cite{Rind2000}, marking a significant milestone in the development of LGMD-based robotics.

    \begin{figure}[t]
		\centering
		\includegraphics[scale = 0.35]{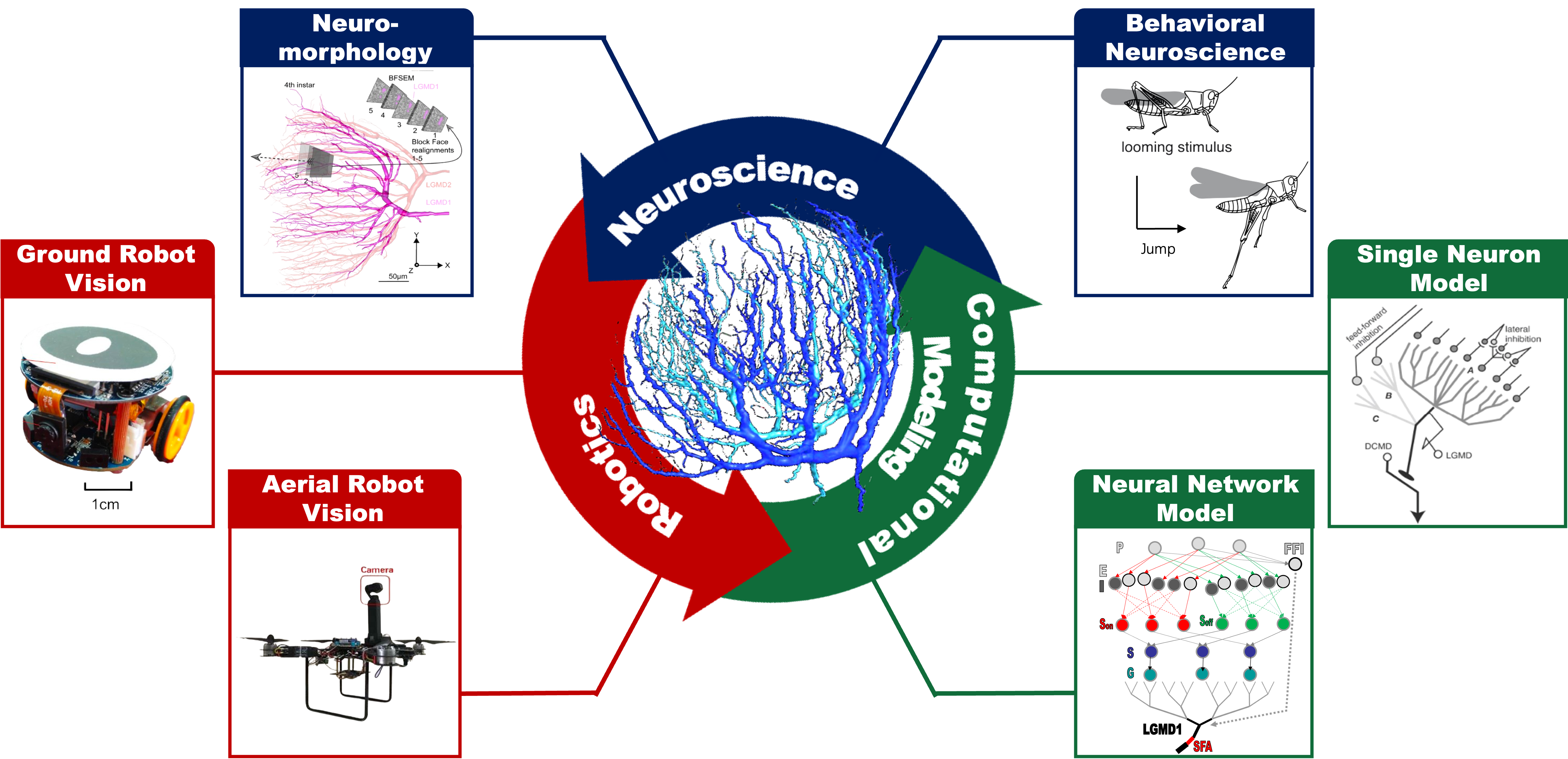}
		\caption{
            The reviewed bio-inspired research paradigm: neuroscience studies of the LGMD neuron, ranging from behavioral neuroscience and neuro-morphology, inform the development of computational models, such as single neuron modeling and multi-layered neural network, simulating LGMD's functionality and selectivity. 
            These models are successfully implemented in robotics as embedded vision, such as micro-mobile robots and small quadcopters, for real-time collision detection and avoidance in navigation. 
            The output and behavior of these LGMD-based embodiments, in turn, validate the models and provide meaningful feedback that inspires further neuroscience research, creating a continuous circle of progress. 
            The insets are adapted from \cite{Rind2014a,Gabbiani2005a,Fu2018,Fu2020b,Zhao2023,Dewell2022,Rind2022}.
            }
		\label{Research_Diagram}
	\end{figure}
    
    As our understanding of LGMDs has expanded to include its morphology, synaptic connectivity, and subcellular localization, both neuroscientists and computational researchers have sought to uncover the intrinsic mechanisms or biological substrates underlying the neuronal characteristics of LGMDs \cite{Gabbiani2005a, Gabbiani2005b, Gabbiani2009a, Dewell2018, Dewell2022, Badia2010, Olson2021}. 
    This growing knowledge has inspired engineers to enhance the performance of LGMD-based robotics by incorporating biologically plausible mechanisms, such as feed-forward inhibition, ON/OFF pathways, and spike frequency adaptation \cite{Fu2018, Fu2019, Fu2023}. 
    These engineering advances not only enhance robotic systems, but also offer valuable insights back into neural dynamics for biologists and computational modelers, demonstrating the functional capabilities and potentials of LGMD visual pathways.

    Neuroscience research on the LGMD neuron has driven the development of computational models and LGMD-based real-world applications, which, in turn, have enriched our understanding of neural dynamics and visual pathway functionality in locusts (see Fig.\ref{Research_Diagram}). 
    Today, this iterative research process has evolved into a mature and continually advancing paradigm, with hundreds of researchers actively contributing to and promoting this field. 
    In this review, we illustrate this bio-inspired research paradigm by summarizing recent advancements in neuroscience, computational modeling, and robotics within the context of LGMD studies. 
    Understanding this paradigm can deepen our insights into the interconnections among these fields and may inspire similar bio-inspired research pathways in other areas of cross-disciplinary research.
	
\section{How LGMD works as looming detectors}
    This section provides an overview of classic biological and computational theories of the LGMD, as well as opening questions regarding the circuits and mechanisms underlying LGMD neural processing. 
    We begin by surveying the classic physiological findings and computational theories including two types of LGMD1 models. 
    These models seek to emulate the intrinsic characteristics of the LGMD1 neuron, either by focusing on dendritic computations within a single neuron or by modeling the LGMD and its afferent signal processing pathway as an integrated functional neural system.
    
    We then elucidate ongoing debates and recent findings that challenge traditional understandings of LGMD neural processing, including alternative hypotheses for neural circuitry, mechanisms of signal integration, and the roles of specific neuronal components. 
    These controversies underscore the complexity of LGMD functioning and motivate continued research in this field.
    
\subsection{Biological briefs}
\begin{figure}[H]
		\vspace{-20pt}
		\centering
		\includegraphics[scale = 0.4]{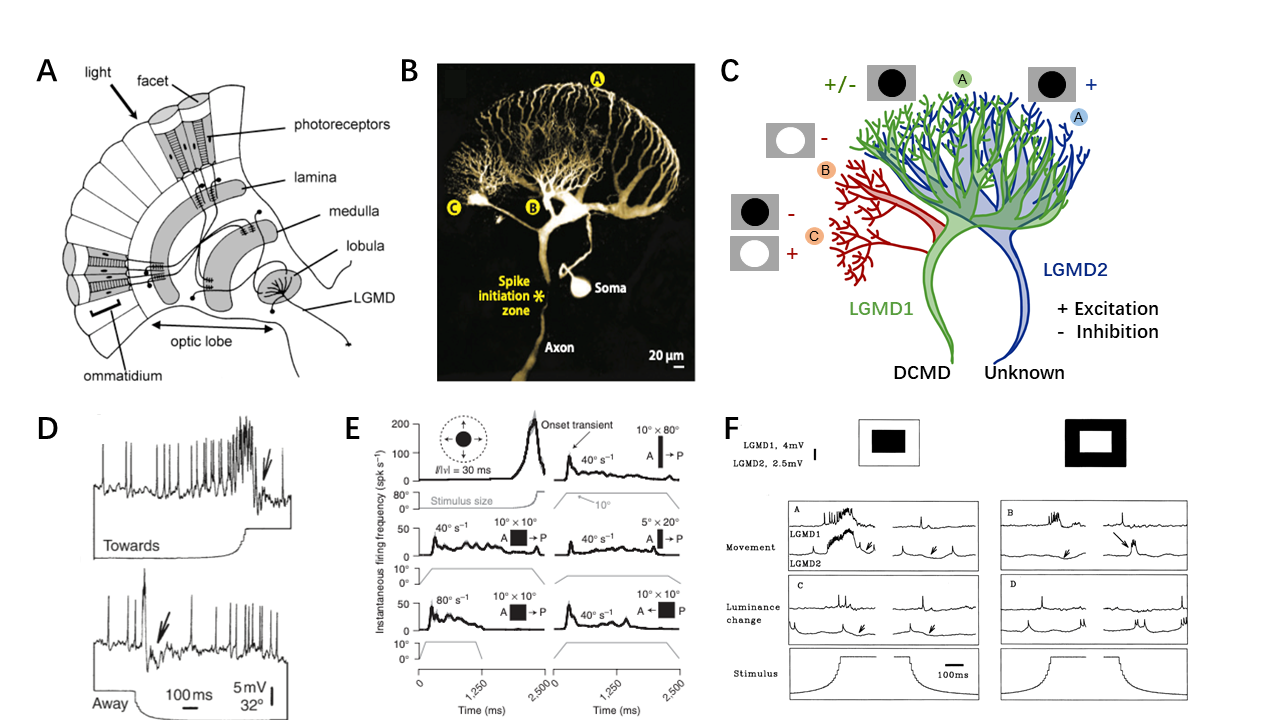}
		\caption{
        The morphology of LGMD1 and LGMD2 with response to different visual stimuli. 
        \textbf{(A)} The afferent network of LGMD1 (adapted from \cite{Rind2002}). LGMD1 is located in the fourth layer of the optic lobe in locusts, with the retina, lamina, and medulla serving as its afferent layers. 
        Specifically, there are $\sim 10$ neurons per lamina column with L1 and L2 proposed to be upstream of the LGMD.
        \textbf{(B)} The morphology of LGMD1 (adapted from \cite{Gabbiani2011}). LGMD1 has three distinct dendritic fields where field A receives approximately $\sim 15,000$ excitatory retinotopic inputs from the entire visual hemifield, while fields B and C receive approximately $\sim 500$ non-retinotopic feedforward inhibitory inputs related to ON and OFF contrasts, respectively. 
        Recent research has shown that dendritic field C also contributes to processing excitatory ON-contrast signals. 
        \textbf{(C)} Schematic diagram of LGMD1 and LGMD2 morphology. The LGMD2 neuron (illustrated in blue) has only a single large dendrite field, and its downstream signaling pathway and postsynaptic targets remain unclear. In contrast, the downstream signaling pathway of LGMD1 has been identified as the DCMD. Dendrite field A (depicted in green) receives excitatory inputs and encodes angular velocity in the OFF contrast. The ``-" symbol indicates that lateral inhibition may also occur within or preceding the dendrite of LGMD1. Dendrite fields B and C (shown in red) receive inhibitory signals from the ON and OFF pathways, respectively. 
        Additionally, recent studies suggest that dendrite field C also processes non-retinotopic excitatory signals from the ON pathway \cite{Dewell2022}. 
		\textbf{(D)} The response of LGMD1 when a locust views a square projected on a screen approaching and then receding at $5 m/s$ (adapted from \cite{Rind1999}). LGMD1 shows a strong and continuous spike train as the square approaches, whereas the neuron displays only a phasic spike at the onset of receding. 
        \textbf{(E)} The response of LGMD1 to approaching and translating stimuli (adapted from \cite{Gabbiani2009a}) - regardless of the size and speed of the translating bar, the instantaneous firing rate of LGMD1 is significantly lower when the locust views translating stimuli compared to looming stimuli. 
        \textbf{(F)} The neuronal responses of LGMD1 and LGMD2 when the locust views light and dark rectangles approaching and receding, as well as in response to changes in luminance (adapted from \cite{Rind1997b}) - LGMD2 is selectively excited by darker objects approaching. 
        The long arrow indicates LGMD2 excitation after the onset of light object receding, while the short arrow indicates hyperpolarization of LGMD2 when both light and dark objects ceased during approach or at the start of receding of a dark object.
        } 
		\label{LGMD_morphology_response}
	\end{figure}

    The LGMD1 was first identified as part of a group of neurons situated in the lobula of the optic lobe in locusts, with its afferent network comprising the retina, lamina, and medulla \cite{Oshea1974} (see Fig. \ref{LGMD_morphology_response}A). 
    Each LGMD1 neuron transmits spikes directly to its postsynaptic target neuron, i.e., the descending contralateral movement detector (DCMD) with a consistent latency. 
    The DCMD connects to the contralateral nerve cord and triggers escape behaviors such as jumping and flight steering. 
    The LGMD1 has three distinct dendritic fields: the largest, dendritic field A, receives excitatory input, while the smaller dendritic fields B and C receive feedforward inhibitory input (see Fig. \ref{LGMD_morphology_response}B).
	
	LGMD1 was initially identified as responding to objects translating in any direction \cite{Oshea1976d}. 
    When tested using a 2D video of an approaching square or circle - commonly referred to as a looming stimulus, LGMD1 exhibited its strongest response to approaching objects under both luminance increase (ON contrast) and luminance decrease (OFF contrast), while responding only briefly during object recession (see Fig. \ref{LGMD_morphology_response}D, E). 
    This looming selectivity remains consistent across various complex movement trajectories and backgrounds \cite{Schlotterer1977,Rind1992,Yakubowski2016}. 
    During 1990s, Rind and Simmons discovered another neighboring neuron to LGMD1 in the lobula, named LGMD2, which shared the similar morphology yet featuring a single fan-shaped dendritic field covering the convex, distal face of the lobula (see Fig. \ref{LGMD_morphology_response}C). 
    Unlike LGMD1, LGMD2 only responds to looming stimuli under OFF contrast \cite{Rind1997b}, such as a dark object approaching against a light background, or a light object receding against a dark background (see Fig. \ref{LGMD_morphology_response}F).
	
\subsection{Classic computational theory on LGMD}
    Regarding computational modeling of LGMDs, the models of LGMD1 emerged since the 1990s \cite{Rind1996a}, while the first model of LGMD2 was proposed over two decades later \cite{Fu2015,Fu2020}. 
    In general, bio-inspired looming perception visual systems can be categorized into two primary classes. 
    The first class focuses on mapping nonlinear relationships between the physical attributes of stimuli (image size and image velocity) and corresponding neuronal activity. 
    The second class utilizes hierarchical neural networks modeled after well-established anatomical pathways responsible for looming perception.
    
\begin{figure}[t]
		\centering
		\includegraphics[scale = 0.5]{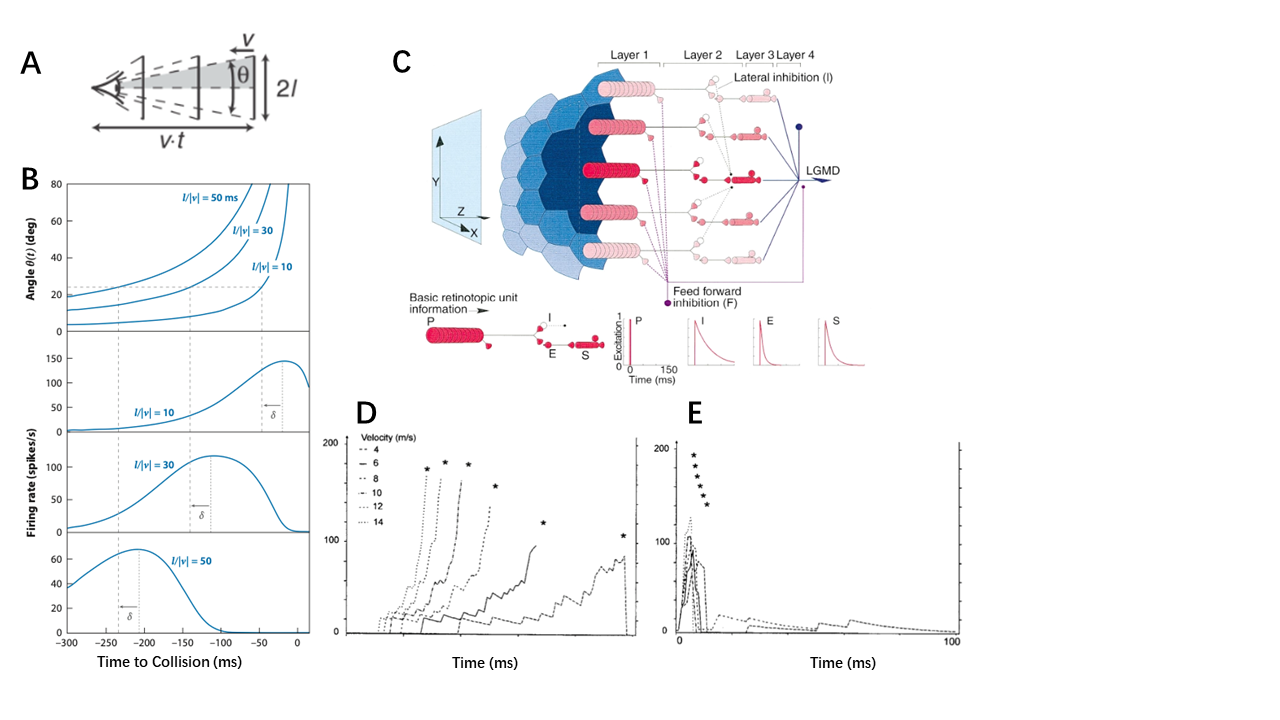}
		\caption{
            Different computational models of LGMD1 and model response.  
		\textbf{(A)} Illustration of a looming stimulus (adapted from \cite{Gabbiani2012}). the diameter of the approaching object is $2l$, and its constant approach velocity is $v$. 
            The angular size $\theta(t)$ projected on the retina can be calculated by $\tan(\theta) = \frac{2l}{v \cdot t}$. and the angular velocity can be determined by taking the derivative of $\theta(t)$. 
            \textbf{(B)} Variation of the $\eta$-function with respect to the size-to-speed ratio $l/|v|$ (adapted from \cite{Gabbiani2011}). as the ratio increases, the approach velocity decreases, leading to an earlier peak in the $\eta$-function. 
            However, the peak neuronal response consistently occurs at a fixed delay of $\delta = 27 \text{ ms}$ after the simulated approaching object reaches a threshold angular size of $24\degree$. 
            \textbf{(C)} The first four-layered LGMD1 neural network proposed by Rind et al. (adapted from \cite{Rind1999}). Unlike the nonlinear mathematical model, Rind and Simmons assumed that the looming selectivity of LGMD1 is shaped by critical image cues and the critical race between excitation and inhibition within its signal processing pathway. 
            \textbf{(D)} and \textbf{(E)} The LGMD1 network output for an approaching and receding square (adapted from \cite{Rind1999}). the asterisk indicates activation of the FFI unit. 
            The four-layered network exhibits an increasing response to looming stimuli, with the peak time of the output occurring earlier for higher approaching velocities. 
            The network also shows a phasic response to object receding, where higher receding velocities correspond to shorter response durations.
            }
		\label{LGMD_models} 
	\end{figure}

	Assuming that dendritic field A of the LGMD1 neuron encodes angular velocity, while dendritic fields B and C encode angular size, Hatsopoulos et al. proposed the first computational model of LGMD1, known as the $\eta$-function \cite{Gabbiani1995}. 
    As illustrated in Fig. \ref{LGMD_models}A, for a looming object with half-size $l$ and constant approach speed $v$, the angular size $\theta$ projected on the retina can be described by $\tan(\theta) = \frac{2l}{v \cdot t}$, which increases rapidly as collision becomes imminent. 
    The angular velocity $\theta'(t)$ is the derivative of angular size with respect to time. 
    The neuronal response of LGMD1 to looming stimuli, from a single neuron computational perspective, is modeled by the following $\eta$-function\footnote{Responses of looming sensitive neurons in both vertebrates and invertebrates could also encode other optical variables like $\tau$ - relative rate of expansion, $\rho$ - absolute rate of expansion \cite{Rind2024}.}:
	$$ \eta(t) = C\theta '(t-\delta)exp(-\alpha\theta(t -\delta)),$$ 
	where the constant $C$ converts angular velocity into the firing rate, while $\alpha$ is a coefficient related to the threshold angular size. 
    The neural delay $\delta$ can be fitted to experimental data. 
    As illustrated in Fig. \ref{LGMD_models}B, the peak neuronal response computed by the $\eta$-function consistently occurs at a fixed delay after the simulated approaching object reaches a threshold angular size on the retina, and this threshold angular size remains invariant with respect to the size-to-speed ratio $l/|v|$ \cite{Gabbiani1999,Gabbiani2001,Gabbiani2002,Gabbiani2004,Gabbiani2011}.

	Researchers have demonstrated that the multiplicative combination of neuronal signals encoding an object's angular size and angular speed can be achieved through a logarithmic-exponential transformation in the LGMD1 neuron \cite{Gabbiani2012}, providing further evidence that the mathematically straightforward $\eta$-function can serves as the intrinsic computational mechanism for the neuronal response of LGMD1. 
    Initially, the $\eta$-function models behave like the LGMD1 only for objects approaching at a constant speed \cite{Gabbiani2023}, lacking the capacity to simulate detailed responses to accelerating objects and other types of visual stimuli, such as recession and translation. 
    Recently, Dewell et al. demonstrated that LGMD responses to approaching objects with non-constant velocities are accurately predicted by the $\eta$ model, exhibiting only minor timing discrepancies \cite{Dewell2023}.
	
	On the other hand, Rind and Simmons proposed a four-layered neural network to model the LGMD1's neuronal response to looming, receding, and translating stimuli \cite{Rind1996a} (see Fig. \ref{LGMD_models}C). 
    In this network, critical image cues (denoted as P in Fig. \ref{LGMD_models}C) are generated by edges that change in extent or velocity during object approach, resulting in strong excitation of computational units (denoted as E in Fig. \ref{LGMD_models}C). 
    Lateral inhibition within these units is propagated with a time delay (denoted as I in Fig. \ref{LGMD_models}C). 
    An additional unit, termed the F unit (denoted as F in Fig. \ref{LGMD_models}C), is integrated into the network to simulate the feed-forward inhibition (FFI) observed in the visual pathway to LGMD1. 
    The captured critical image cues, the critical race between excitation and inhibition and the FFI together fine tune the looming selectivity of the proposed neural network, showing strong and increasing response to looming stimuli, while transient response to object recession and minor response for single edge crossing (elongation) (see Fig. \ref{LGMD_models}D, E). 
    Reducing the strength of lateral inhibition significantly raises the network's response to object receding and translating. 
    Moreover, removing the FFI prolongs the network's response to both object approach and recession.
	
	Unlike computational models as the $\eta$-function, the hierarchical four-layered LGMD1 neural network can process video streams as input, which can be easily accessed through CCD or CMOS cameras. 
    With strong looming selectivity compared to typical visual stimuli, such as receding and translating, the LGMD1 neural network made its first attempt to interact with real-world environments in 2000, successfully achieving collision detection and avoidance for a mobile robot \cite{Rind2000}. 
    Recent advances of biological studies on LGMD neurons have significantly enhanced the four-layered LGMD1 neural network, improving its ability to handle widely existing, real-world challenges. 
    By integrating additional mechanisms discovered in the visual pathway to LGMD1, such as ON/OFF competition \cite{Fu2020b,Lei2023} and spike frequency adaptation \cite{Fu2018}, the performance of LGMD1-based artificial systems have been substantially improved. 
    These enhancements have been successfully transformed to vision systems in mobile robotics and umanned aerial vehicles (UAVs), demonstrating the merit of LGMD-inspired neural networks for collision detection in complex and dynamic environments, reminiscent of LGMDs in locusts.
	
\subsection{Controversial arguments on LGMD neural processing}
	Although both categories of computational models effectively capture the neuronal characteristics of the LGMD1 when faced with looming objects, a key point of contention between these models lies in the \textit{timing} of the peak response. 
    In the first class of models, the $\eta$-function produces a peak response after a fixed delay, following the point at which the looming object reaches a specific image size threshold. 
    However, the peak will be occurring after the object projects its maximum size onto the retina when $l/|v| \leq 5\text{ ms}$ \cite{Gabbiani1999b}.
    This ratio can be achieved by, for example, an 8 cm-wide object moving at $8$\text{ m/s} ($28.8$\text{ km/h}) or a 4 cm-wide object traveling at $4$\text{ m/s} ($14.4$\text{ km/h}). 
    In contrast, some experimental results for the second category of models, as reviewed in \cite{Fu2019}, indicate that the peak response consistently occurs before the looming object reaches its maximum size on the RF.
	
	While the hierarchical neural network proposed by Rind et al. clearly demonstrates how looming selectivity is generated in the LGMD1 neuron, it is worth noting that the proposed four layers may not anatomically correspond to the four layers of the locust's stratified optic lobe.
	
	Firstly, the photoreceptors in the retina primarily convert luminance changes into electrical signals while preserving the retinotopic organization \cite{Lilywhite1978, Wilson1982, Williams1982, Kuster1985a, Kuster1985b}. 
    However, it remains largely unknown how these photoreceptors capture or derive the critical image cues required for LGMD responses. 
    It is also possible that these cues are derived from other layers or from the combined processing of multiple layers in the visual neural pathway. 
    Additionally, the retina may play a role in light and dark adaptation, contributing to the LGMD neuron's ability to respond accurately under varying background-object contrast conditions \cite{Tunstall1967, Wilson1975}.
	
	Secondly, while ON and OFF contrasts are separated in lamina monopolar cells (LMCs) located in the lamina, the precise mechanisms of how these signals combine or interact, particularly in the optic lobe of locusts, are unverified. 
    Earlier research on LGMD1 suggested that ON and OFF contrast signals jointly flow into the medulla layer, where they transmit to dendritic field A for excitation, and to dendritic trees B and C for ON/OFF-contrast FFI, respectively \cite{Gabbiani2004}. 
    However, recent neuroscience studies indicate that dendritic field C may partially contribute to excitation in response to ON contrast stimuli by receiving inputs from neurons in the dorsal uncrossed bundle (DUB), which possess smaller receptive fields ($\sim 10 \degree$). In contrast, OFF contrast-sensitive neurons responsible for FFI are fewer in number and exhibit broader receptive fields ($\sim 50 \degree$) \cite{Dewell2018,Dewell2022,Wang2018}. 
    Moreover, it has been mentioned that global inhibition, scaled by overall laminar activity, functions as a normalization mechanism in the medulla layer for increasing excitation as collision approaches \cite{Zhu2018}. 
    Computational modelers have attempted to replicate this function by introducing a non-spiking neuron in the lamina layer \cite{Olson2021}. 
    Such GI neuron gather signals across LMCs then affect the trans-medullary afferent neurons (TmAs) in the medulla where the authors suggested it may function as a network of neurons. 
    In addition, they also modeled lateral inhibition crossing between units directly from the LMCs \cite{Olson2021}.  
    However, anatomical evidence may conflict with this modeling assumption, as each ommatidium has its own set of photoreceptors and LMCs, and LMCs are believed to interact only within their own set and not across different ommatidia \cite{Rind2014}.
	
	Thirdly, regarding the TmAs as inputs to LGMDs (targeting dendritic field A of both LGMD1 and LGMD2), early neural modeling efforts assumed that TmAs were responsible for lateral inhibition in the LGMD pathway \cite{Badia2010, Olson2021}. 
    When traced for reconstruction, TmAs indeed exhibited lateral connections and received input from the outer medulla layer. 
    Most works demonstrated such lateral interactions are likely inhibitory types \cite{Rind1998,Rind2014b,Rind2022}. 
    However, evidence suggests that this lateral interaction contributes to excitation rather than inhibition\cite{Zhu2018}, and the lateral excitation involving TmAs is thought to enhance LGMD's responses to coherent stimuli \cite{Zhu2018}.
    Another study posited that hyperpolarization-activated cation (HCN) channels are responsible for increasing neuronal preference for coherent looming stimuli \cite{Dewell2018}. 
    Through two works specializing in dendritic synaptic integration \cite{Greg2015,Hu2018}, we infer that the density of HCN channels may generally increase with dendritic distance from the soma, i.e., distance-based synaptic integration in the LGMD's dendritic field may also contribute to an increased response to coherent stimuli.
    Additionally, transient cells may exhibit strong self-inhibition, which could reduce LGMD responses to translating stimuli \cite{Rind2016}. 
    Together, these findings indicate that the mechanisms underlying lateral and self-inhibition, as well as excitation in LGMD pathways, are far more complex than our current thoughts. 
    This highlights the need for more investigations into the interplay between these components in generating LGMD's neuronal responses.

    \begin{figure}[htp]
    	\vspace{-20pt}
		\centering
		\includegraphics[scale = 0.28]{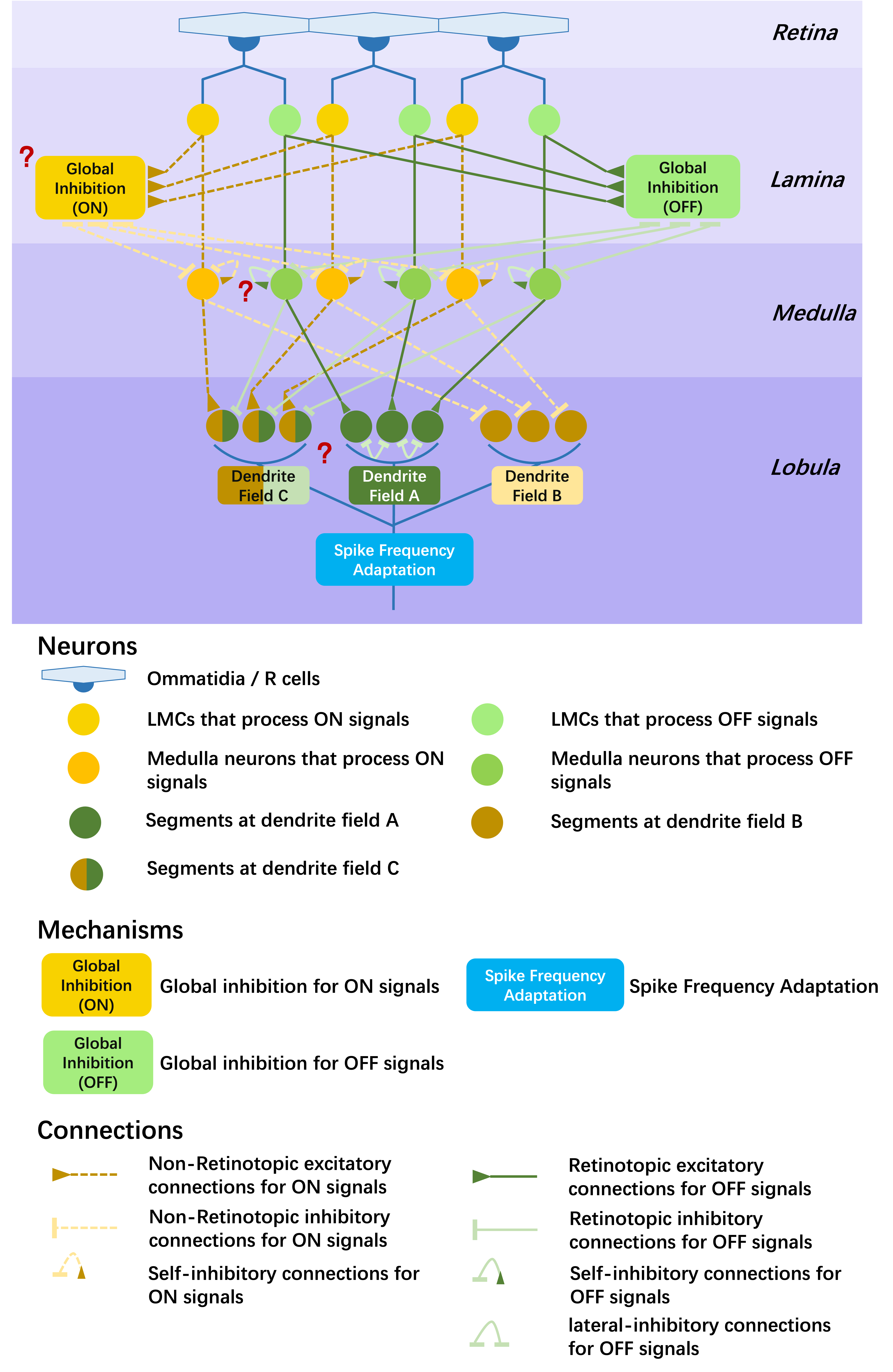}
		\caption{
        The diagram of a possible visual signal processing of LGMD from our perspective and assumption: there are three key areas represented by question marks that highlight the unknown mechanisms in LGMD circuit processing. 
        First, it remains unclear whether a separate neuron is responsible for gathering global information and subsequently inhibiting other neurons. 
        To date, the exact location for global inhibition is still mysterious. 
        In line with a recent modeling work \cite{Olson2021}, we assume that this inhibition is mediated by a separate neuron in the lamina layer. 
        Second, the mechanism of lateral excitation remains poorly understood. 
        Research has indicated that lateral excitation operates on a larger time scale compared to lateral inhibition \cite{Zhu2018}, and most likely occurs within trans-medullary afferent neurons (TmAs) \cite{Rind2016}. 
        In the processing diagram presented here, we have not explicitly depicted lateral excitation since the neurons in the medulla are classified by ON/OFF contrasts, rather than as specific neurons like TmAs or DUB neurons. 
        Moreover, we assume that ON and OFF signals at the medulla layer are completely separate, as how these signals interact is still unknown. 
        Third, it is uncertain whether interactions occur within the dendritic trees or how integration occurs near the spike initiation zone (SIZ). 
        In this diagram, we assume that the signals from the three dendritic trees are linearly summed without interaction, and then transmitted to the SIZ. 
        After passing through the spike frequency adaptation, the combined signal is subsequently conveyed to the DCMD towards motor system.
        }
		\label{Overall_pathway} 
	\end{figure}
	
	Fourthly, dendritic fields B and C of the LGMD1 neuron exhibit phasic ON and OFF inhibition as part of the FFI mechanism, which increases in parallel with the excitation observed in dendritic field A. 
	The FFI signal is dependent on the size of the looming object rather than its approach velocity \cite{Gabbiani2005a,Wang2018}. 
    Earlier models assumed that FFI was directly received by dendritic fields B and C from TmAs, the same input neurons providing excitation to dendritic field A. 
    However, recent studies have suggested that FFI, delivered to field C, is derived instead from DUB neurons. 
    The receptive fields of DUB neurons are highly overlapped, and these neurons show no directional selectivity. 
    In general, DUB neuron firing rates persist for at least $100$\text{ ms} after the LGMD stops responding and then gradually return to the resting level \cite{Wang2018}. 
    Despite these insights, the existence and nature of interactions between DUB neurons (which provide inhibitory input to the LGMD) and TmAs (which provide excitatory input) remain elusive. 
    This gap in knowledge underscores the complexity of inhibitory and excitatory interactions within the LGMD's afferent circuitry.
	
	From our current understanding, the potential visual signal processing pathway for LGMD can be hypothesized as illustrated in Fig. \ref{Overall_pathway}. 
    More specifically, for a looming stimulus, the R cells or photoreceptors within each ommatidium convert the captured luminance changes into electrical signals and pass them to their respective sets of LMCs, where ON/OFF signal separation occurs. 
    Furthermore, global inhibition, which can serve to normalize excitation input to a reasonable level, may occur within each LMC or through a separate pathway. 
    The separated ON/OFF signals are then passed to the medulla layer, where they are processed by TmAs and DUB neurons, respectively. 
    Within the TmAs, self-inhibition reduces the response of the LGMD to translating objects. 
    The TmAs could also exhibit lateral excitation, which enhances LGMD responses to coherent stimuli. 
    The excitatory OFF contrast signals processed by TmAs are then transmitted to dendritic field A of the LGMD, with a portion of the ON contrast excitation signal also passing to dendritic field C. 
    DUB neurons, characterized by relatively large receptive fields, encode the angular size of looming objects in both ON and OFF contrasts and transmit these signals to dendritic fields B and C, respectively. 
    Lateral inhibition occurs at the surface of the LGMD, and synaptic integration can also occur within each dendrite, based on the distance between the dendrite and the soma. 
    Signals from the three dendritic fields are combined at the SIZ, where spike frequency adaptation occurs. This mechanism can be considered a form of habituation, reducing the neuronal response to translating stimuli while enhancing the response to looming stimuli.
    Finally, the spikes generated by the LGMD are transmitted to its postsynaptic partner DCMD, triggering escape behavior. 
    In general, this hypothesized model could replicate biological functionality, though certain implementation details do not precisely reflect biological realism.
	
\section{LGMD-based dynamic vision systems facilitate collision-free navigation}
\subsection{LGMD1-based collision detectors for ground and aerial robots}
	Real-world applications based on LGMD neurons began to emerge with the advent of LGMD computational models. 
    As discussed previously, the first type of LGMD1 model can be described by the $\eta$ function, which integrates inhibitory signals that encode the angular size ($\theta$) with excitatory signals encoding angular velocity ($\dot{\theta}$) through an exponential formulation. 
    In addition to the $\eta$ function, Keil introduced a new function called the $\psi$ function, which follows the dynamics of the RC circuit. 
    While the $\psi$-function is also capable of fitting experimental data that aligns with the $\eta$-function, it differs by avoiding the biophysical challenges associated with implementing exponential inhibition \cite{Keil2011}. 
    Furthermore, the $\psi$-function was enhanced by incorporating biologically plausible inhibition regulation, resulting in a model that generates a response similar to the $\eta$-function, even in the presence of noisy information channels \cite{Keil2015}.
	
	Although these models are established with biologically plausible formulations, they are rarely used to address real-world challenges directly owing to their reliance on the physical attributes of looming objects that are impossible to measure in unpredictable, real-world environments. 
    To alleviate this, Keil developed a multi-layer network that uses luminance values, captured by CCD or CMOS cameras, as model input. 
    The response of this model exhibits characteristics similar to those of the $\eta$-function, without prior knowledge on image size or velocity \cite{Keil2003}. 
    This model was subsequently extended with ON/OFF pathways to enhance its performance across real-world scenes \cite{Keil2004}. 
    However, the computing units in these models are formulated using ordinary differential equations, which makes the computations time-consuming and limits their applicability in robotics for real-time visual processing and navigational applications. 
    Additionally, these approaches employ hierarchical structures similar to the four-layer LGMD1 network proposed by Rind, rather than the single-neuron computation represented by the $\eta$-function.

    \begin{figure}[t]
		\centering
		\includegraphics[scale = 0.28]{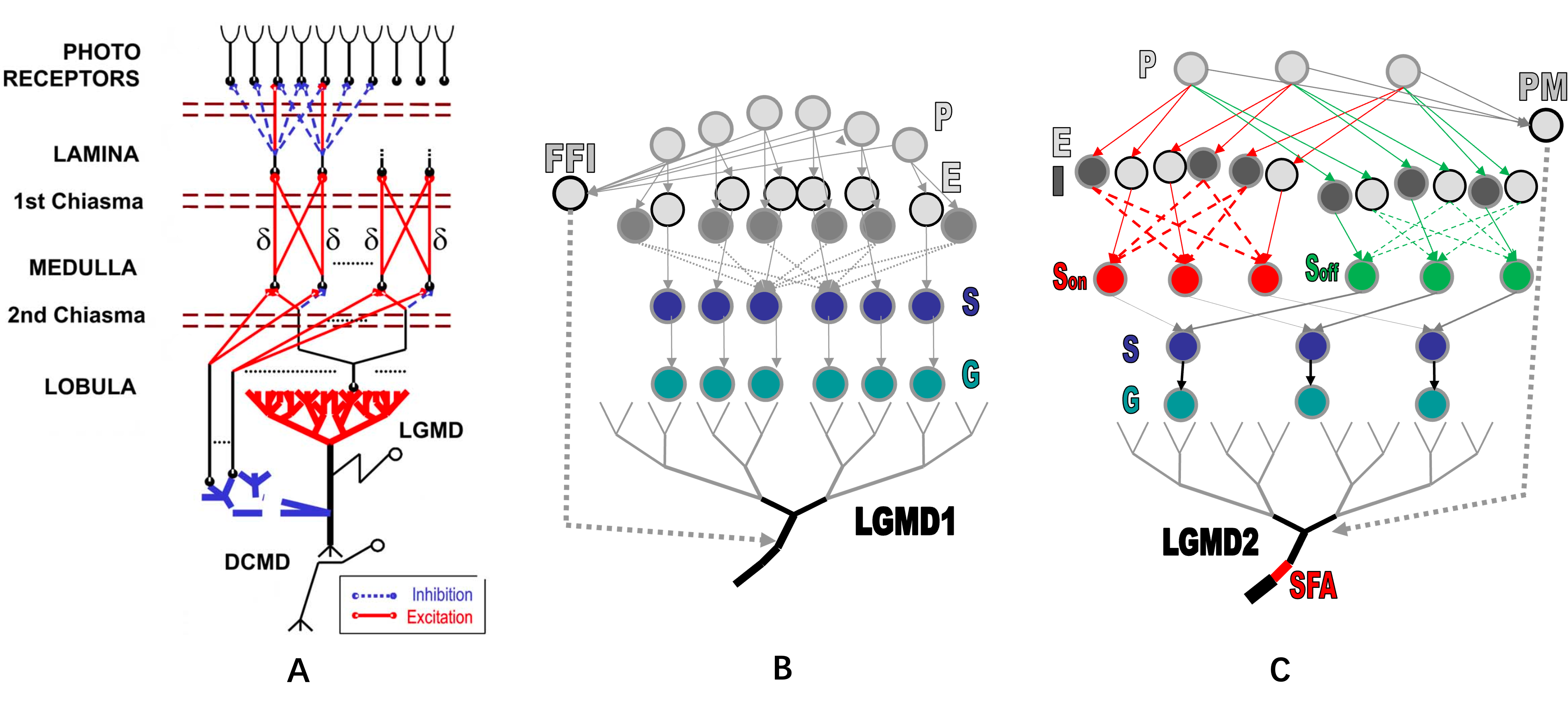}
		\caption{
            The hierarchical structures of three typical LGMD neural network models. 
            \textbf{(A)} The four-layered LGMD network proposed by Badia et al. (image courtesy of \cite{Badia2010}). This model integrates the activity of a set of ON/OFF-type neurons from the medulla that surround each location. 
            These neurons also provide input for generating FFI. 
            \textbf{(B)} The four-layered LGMD1 network proposed by Yue and Rind. (image courtesy of \cite{Yue2006a}). This model introduces a artificial grouping-layer to enhance the previous LGMD network by Rind et al. \cite{Rind1996a}. 
            This improves the model's response to expanding edges while reducing sensitivity to isolated noise when coping with real-world visual stimuli. 
            \textbf{(C)} The LGMD2 neural network proposed by Fu et al. (image courtesy of \cite{Fu2020}). This network incorporates ON/OFF pathways and spike frequency adaptation mechanism in this framework, which significantly improve the looming selectivity. The LGMD2 neuron has only one dendrite, which shares a similar shape with dendrite field A of the LGMD1 model. Consequently, the feedforward inhibition (FFI) mechanisms present in dendrite fields B and C of the LGMD1 model are absent. To compensate for this, an adaptive inhibition mechanism is introduced to suppress the model’s response to whole-field motion.
			Although the frameworks presented in Figures B and C do not strictly correspond to the anatomical structure of LGMD neurons, they are widely used in LGMD-based robotic applications, supporting both ground mobile robots and UAVs for highly efficient collision-free navigation.
            }
		\label{4-layer LGMD model} 
	\end{figure}
	
\begin{figure}[htp]
		\centering
		\includegraphics[scale = 0.35]{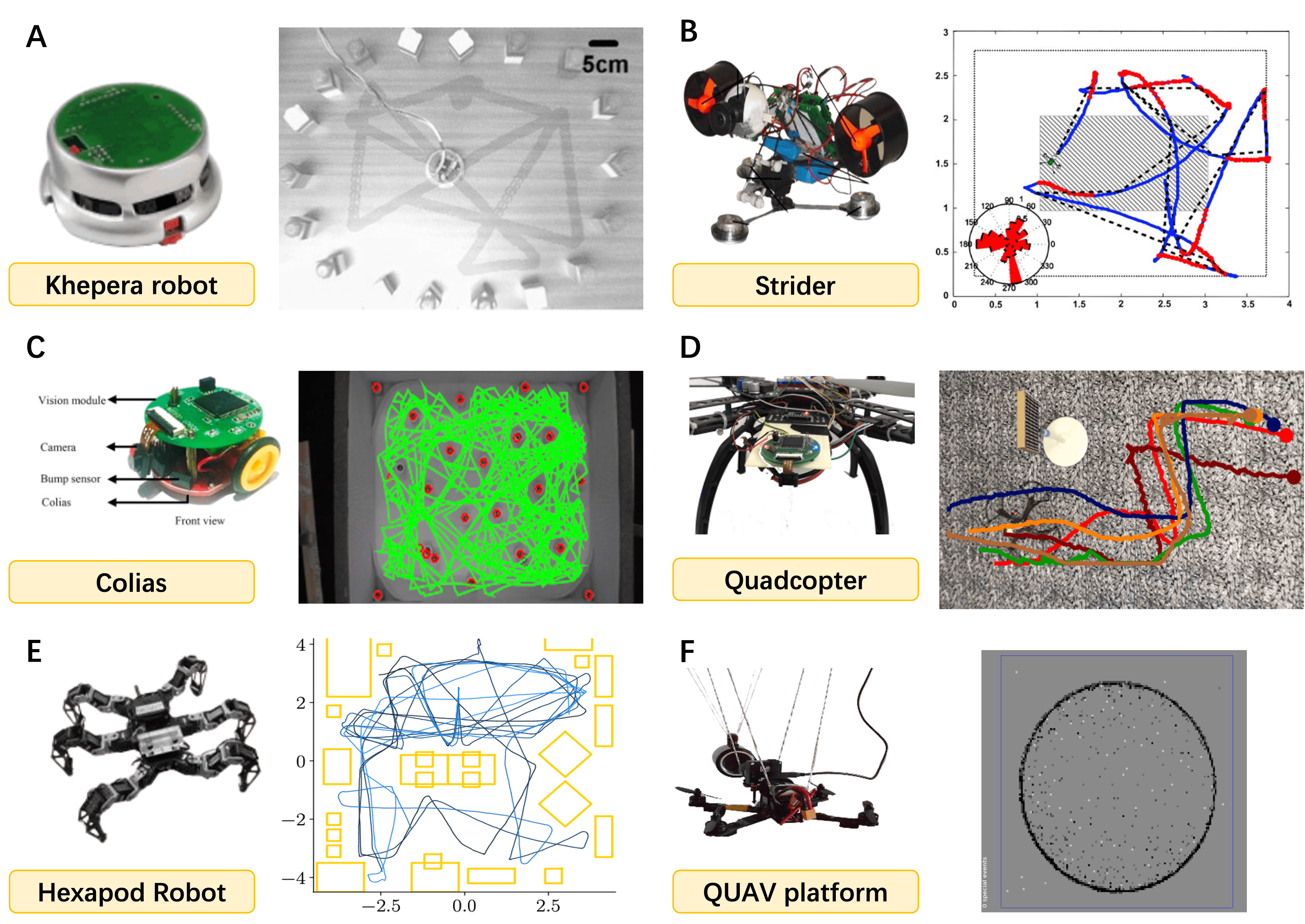}
		\caption{
            LGMD-based collision detectors for ground and aerial robots. 
            \textbf{(A)} Khepera robot and its moving trajectory (image courtesy of \cite{Yue2006a}). A CCD camera was mounted on the top of robot, which was connected to a simulation system. 
            Controlled by the LGMD-based collision detection system, the robot can navigate freely even when surrounded by obstacles. 
            The trajectory in panel-A depicts the robot's path when moving at a speed of $9.6 \text{ cm/s}$. 
            \textbf{(B)} ``Strider" and its navigation trace in a $3 \times 4 \text{ m}^2$ arena (image courtesy of \cite{Badia2010}). The robot is guided by an LGMD model in conjunction with an elementary motion detector (EMD)-based course stabilization system. 
            In panel B, blue traces represent the robot's trajectory, whereas red segments highlight the detection of imminent collisions. 
            Black dashed lines are obtained by fitting linear segments to the robot's trace data, minimizing the mean square error. 
            The inserted panel provides a depiction of the robot's heading direction. 
            \textbf{(C)} \textit{Colias} and its navigation trace in a $3 \times 4 \text{ m}^2$ arena (image courtesy of \cite{Hu2017}). The \textit{Colias} robot features a vision module mounted on top of the platform, with an LGMD-based collision detection system embedded into its onboard chips. 
            The trajectory illustrates \textit{Colias} navigating through an arena containing $18$ obstacles, with a moving speed of $17.9 \text{ cm/s}$ over time.
            \textbf{(D)} A small quadcopter equipped with an embedded LGMD vision detector and its movement trajectory (image courtesy of \cite{Zhao2018}). The obstacle is a patterned board, and the trajectories of the vehicle's center point are depicted on the image in different colors, with the starting point marked by dots. 
            These trajectories clearly demonstrate that the quadcopter successfully avoids the obstacle during each flight. 
            \textbf{(E)} A hexapod walking robot and its trajectories in an indoor environment over $5 \times 8 \text{ m}^2$(image courtesy of \cite{Cizek2019}). The robot utilize the LGMD neural network for visual interception detection and a central pattern generator for locomotion control, as well as a long short-term memory recurrent neural network. 
            The yellow squares represent the obstacles, and the blue lines indicate the five trajectories of the robot, in which the total traveled distance by the robot has been measured to $235.3 m$, showing great feasibility of the LGMD-based hexapod walking robot under the crowded real-world environment. 
            \textbf{(F)} An UAV platform and its output from a dynamic vision sensor (DVS) for a looming circle stimuli (image courtesy of \cite{Salt2017}). The adapted neuromorphic processor uses an LGMD-DCMD model and the output of a DVS, allowing a low-power implementation of the collision detection system, suitable for real-time navigation control.
            }
		\label{LGMD-robotics} 
	\end{figure}

	In contrast, the four-layer LGMD neural network proposed by Rind et al. appears particularly attractive to robotic engineers. 
    The first implementation of this four-layer LGMD network for collision detection in ground mobile robots occurred in 2000, where a Khepera mobile robot equipped with a CCD camera was used to perceive real-world environments. 
    Collision detection and avoidance were achieved using the simulation software IQR421 \cite{Rind2000}. 
    Subsequently, Yue et al. combined genetic algorithms with the LGMD neural network to address collision detection in car driving scenarios \cite{Yue2006b}. 
    Further improvements were made by incorporating artificial pathway and layer into the four-layer LGMD network, thereby enhancing its performance when managing complex and dynamic environments, alleviating negative impact by camera exposure \cite{Yue2006a} (See Fig. \ref{4-layer LGMD model}B and Fig. \ref{LGMD-robotics}A). 
    Building on similar ideas, Yue and Rind continued exploring the potential of LGMD-based near-range path navigation, including the integration of panoramic images and regular CCD images for improved collision detection and avoidance \cite{Yue2009, Yue2010}. 
    Moreover, Silva et al. extended the LGMD model by integrating previous work from \cite{Yue2006a} and \cite{Meng2009}. 
    They verified a linear relationship between time-to-collision\footnote{time difference between peak firing time of LGMD model (predicted time) and genuine collision time} and time-of-collision, then investigated the $l/|v|$ ratio in MATLAB simulations and in online DKRK8000 robot navigation \cite{Silva2014}.
    
    Another milestone was established in 2017 when the LGMD-based neural network model was built for the first time into the embedded vision system. 
    Hu et al. adapted a similar vision-based LGMD network as in \cite{Yue2006a} into a small autonomous mobile robot called ``\textit{Colias}" \cite{Hu2017}. 
    The robot performance was validated through autonomous navigation in an arena mixed with obstacles (see Fig. \ref{LGMD-robotics}C), demonstrating the reliability and robustness of the embedded vision system of LGMD case. 
    Later, Fu et al. further facilitated collision-free navigation for \textit{Colias} robot by gradually incorporating spike frequency adaptation, ON and OFF visual pathways, as well as feedback neural computation \cite{Fu2018, Chang2023} (See Fig. \ref{4-layer LGMD model}C). 
    In the latest works, the feedback closed-loops in ON/OFF channels are capable of implementing ON/OFF-contrast loom-selectivity of the micro-robot through easily mediating a singular parameter \cite{Chang2023, Chang2024}.
    Except the wheeled robots, a computational study \cite{Cizek2017} also showcased success of the LGMD1 model as collision detector in a hexapod robot walking through a structured indoor environment (see Fig. \ref{LGMD-robotics}E).
    
    Regarding aerial robot scenarios, Zhao et al. utilized LGMD-based vision detectors to achieve collision-free flight of a small quadcopter and the model is highly biomimetic that simulates spatiotemporal interaction between excitation and inhibition throughout the pre-synaptic connections (see Fig. \ref{LGMD-robotics}D)\cite{Zhao2018,Zhao2019,Zhao2023}. 
    More recent studies have demonstrated that a variety of LGMD-based vision system also holds great promise for boosting the safe navigation of small UAVs \cite{Zhao2023,Zhao2024}. 
    The UAV mounted with an attention-based LGMD visual system can even detect and flexibly avoid thin power lines during flight \cite{Zhao-IJCNN-2022}.
    
    Apart from the Rind's hierarchical neural network, there was also other bio-robotic studies showing that relying solely on vision, the LGMD system is sufficient for collision detection in a blimp-based UAV during indoor flights \cite{Badia2004, Badia2007}. 
    By integrating leaky integrate-and-fire and leaky linear-threshold neurons, the behavioral implications of the LGMD model were evaluated using a ball-caster-based robot platform called ``Strider", demonstrating that local nonlinear computations can also be applied to mobile robots for effective and real-time collision detection \cite{Badia2010} (LGMD model in Fig. \ref{4-layer LGMD model}A and robot in Fig. \ref{LGMD-robotics}B).
	
	Furthermore, engineers and computational modelers have explored ways to enhance the looming selectivity of LGMD1-based perception methods by incorporating various computational mechanisms. 
    These methods can be broadly classified into two categories: the first approach involves integrating mechanisms such as ON/OFF competition or directional sub-networks to suppress responses against receding or translating stimuli \cite{Yue2007, Lei2023}. 
    The second approach employs binocular vision or additional units to estimate changes in the depth of moving objects, thereby distinguishing between approaching, receding, and translating stimuli \cite{Meng2009, Xu2023, Zheng2023}. 
    However, these mechanisms could essentially increase computational burden, have not yet been validated by real-world robotic implementations.
	
	\subsection{LGMD2-based collision detector with enhanced selectivity for ground robot}
	On the other hand, due to limited biological research on LGMD2 and its selective response to only OFF contrast looming stimuli, i.e. proximity of darker object relative to background, LGMD2 has relatively very limited numerical modeling and applications in robotics. 
    The seminal work on LGMD2 modeling appeared late in 2015 when Fu and Yue introduced ON and OFF visual pathways and non-linearity to simulate the LGMD2's selectivity for OFF contrast stimuli \cite{Fu2015}. 
    According to its specific selectivity that matches well with the visual events faced by ground robots where most foreground moving targets intrinsically become darker than their background, the LGMD2 neural network model was optimized and smoothly embedded into the \textit{Colias} micro-robot, achieving a collision detection and avoidance success rate of 95.3\% in arena tests, showing improved selectivity to approaching over receding and translating stimuli \cite{Fu2016}.
    
    As LGMD1 and LGMD2 have different selectivity that could benefit addressing variable real-physical collision challenges, the first hybrid LGMD1-LGMD2 neural model was proposed to implemented as embedded vision system for micro-robot navigation through both dark and bright environments \cite{Fu2017}. 
    In this research, the robot's LGMDs vision behaved similarly to revealed parts of biological neurons that demonstrated the effectiveness of LGMD1 and LGMD2 together for artificial collision-detecting system. 
    Unlike LGMD1, LGMD2 possesses a single large dendritic field, leading to the assumption that FFI is absent in its pre-synaptic circuitry. 
    To accurately simulate the LGMD2's selectivity, Fu et al. introduced an adaptive inhibition mechanism as a substitute for FFI, thereby aligning with physiological observations and enhancing the performance of the LGMD2-based visual system in real-world tasks including complex car and UAV scenarios \cite{Fu2020,Fu2020b, Fu2023b}.
	
	\subsection{New application scenarios and sensor techniques with LGMD-based dynamic vision}
	Researchers are actively seeking opportunities to extend LGMD research to new application scenarios and sensor technologies. 
	A hybrid LGMDs visual system, integrating the firing rate coordination of LGMD1 and LGMD2 neuronal models, was developed as a visual sensing modality for individual micro-robots to enhance intelligent traffic scenarios \cite{Fu-Frontiers-LGMDs}. 
	This system aims to address automatic collision detection challenges in city and highway traffic simulations. 
	This research represents an initial attempt to evaluate the effectiveness of LGMD-based visual systems in dynamic traffic environments, tractably and at low price, marking a step toward the development of neuromorphic collision sensors for autonomous vehicles. 
	Benefiting from its low computational power, LGMD-based collision detection methods have been combined with path integration mechanisms discovered in sweat bees and desert ants, enabling collision-free autonomous navigation toward specific goal locations, even in the presence of both static and moving obstacles \cite{Sun2023, Sun2024a, Sun2024b}.
	
	Unlike conventional cameras, event-based cameras detect changes in light intensity asynchronously, producing sparse, high-temporal-resolution data with low power consumption and a wide dynamic range \cite{Gosh2023}. 
	This makes event-based cameras ideal inputs for motion perception models. 
	When interpreted with spiking neural networks (SNN), event camera-based LGMD models compensate for the limitations of traditional RGB cameras, effectively detecting ultra-fast looming objects or targets in complex and diverse low-light environments, even facilitating the control of UAVs during flights \cite{Wang2024, Deng2024,Salt2017}. 
	Recently, a fractional spiking neuron model was proposed to simulate the spiking initiation zone of the LGMD, where ON/OFF events from a dynamic vision sensor are encoded as inputs \cite{Deng2024}. 
	This research validates the effectiveness of this sensor strategy in emulating the neuronal dynamics of LGMD. 
	Furthermore, with optimization techniques, Salt et al. utilized an event-driven sensor to develop an LGMD-based UAV obstacle avoidance algorithm, demonstrating that the internal spiking dynamics can be effectively represented within a spiking LGMD neural network \cite{spikingLGMD-TNNLS}. 
	Moreover, thermal cameras have been integrated into LGMD computational models to enhance performance \cite{Zhang2023}. 
	Unlike previous approaches that relied on normalization mechanisms to boost responses to low-contrast stimuli, Zhang et al. recently introduced thermal imaging as an alternative solution. 
	By leveraging thermal cameras, the ability of LGMD-based models to detect approaching objects was significantly improved, particularly in low-light conditions, offering a candidate method for collision detection in challenging lighting environments.
	
	In summary, LGMD-based collision detection methods provide a distinctive, visually driven, and data-free approach, standing in stark contrast to deep learning and sensor-based techniques. 
    While deep learning models achieve remarkable accuracy by identifying intrinsic patterns within large datasets, they demand significant computational resources, extensive data, additional sensors, and lengthy training processes. 
    In comparison, LGMD-based methods leverage biologically plausible neural structures to enable parsimonious and efficient collision detection without requiring high-performance hardware or large datasets. 
    This makes them particularly well-suited for real-time computing in mobile machines, where timeliness, computational simplicity, and energy efficiency are paramount. 
    Despite these advantages, LGMD-based models face notable challenges in robustness when applied to diverse, variable real-world scenarios. 
    They lack the adaptability and flexibility demonstrated by real LGMD neurons, especially in handling the complexities of dynamic and noisy environments. 
    Addressing these limitations presents significant opportunities for future research to further refine and expand LGMD-based dynamic vision systems and their applications.
	
\section{How these methods feedback to neuroscience} 
    Can bio-inspired modeling studies and bio-robotic approaches positively influence neuroscience research? 
    This open question has sparked considerable interest across various disciplines. From our perspective, the answer is an unequivocal ``yes". 
    Bio-inspired modeling and bio-robotic approaches do not merely emulate biological systems; they actively contribute to understanding the underlying principles of neural and behavioral mechanisms. 
    By replicating and testing biological phenomena in controlled environments, these approaches provide testable hypotheses that can validate, refine, or challenge existing neuroscience theories. 
    Moreover, they offer insights into the neural substrates of behavior, bridging gaps between abstract models and tangible biological processes.
	
    The performance of LGMD-based robotic systems during complex real-world interaction tasks has directly demonstrated a robustness in looming selectivity comparable to that observed in biological LGMD neurons. 
    In particular, the computational model of LGMD2 has been integrated into the micro-robot - \textit{Colias}, demonstrating a selective looming response to darker objects during real-world navigation \cite{Fu2020}. 
    Additionally, the Strider robot \cite{Badia2010}, implemented with a four-layer LGMD1 network incorporating a different correlation framework, revealed a linear relationship between time to collision and the size-to-speed ratio ($l/|v|$), closer to previous findings by Gabbiani et al. \cite{Gabbiani2002, Gabbiani2004}.  
    F. Claire Rind, a neuroscientist who has dedicated several decades to studying the LGMD neurons, remarked that \textit{``Insects, and locust looming detectors in particular, have already provided inspiration for visual control of robots, unmanned autonomous vehicles, and aerial drones."} in a recent discussion on insects' 3D vision \cite{Rind2024}.
    Her insights underscore the profound impact of bio-inspired research in bridging the gap between neuroscience and technological innovation, demonstrating how understanding simple biological systems can drive advancements in robotics and autonomous systems. 
    Actually, the earlier work based on Rind's LGMD1 neural network \cite{Rind1996a},  i.e., the Khepera robot, controlled by an LGMD-based collision detection and avoidance system, exhibited reasonable behavior—specifically, as the moving speed increased, the robot generated collision warnings earlier, thereby maintaining a greater distance from obstacles \cite{Yue2006a}.
	
	Bio-inspired robotics have the potential to contribute far beyond mere ``verification". 
	Two decades ago, scholars proposed that bio-robotic studies could serve as effective paradigms for understanding animal behavior, suggesting that the performance of bio-inspired robotics might also offer valuable insights into neuroscience and behavior \cite{Webb2001}. 
	Regarding LGMD2, the strongest feedback to neuroscience is the probable existence of ON/OFF-contrast encoding neurons or neural pathways prior to the dendrites of LGMDs \cite{Fu2020,Fu2023}. 
	At this point, Rind also commented \textit{``Simulations show how response tuning for dark transitions in the LGMD2 could complement LGMD1 responses"}\cite{Fu2020}. 
	Unlike research into fly visual systems, there is very limited evidence for whether/where the ON/OFF channels exist in locust's visual circuitry. 
	However, the long-term knowledge of separate channels for ON/OFF-contrast FFI in the LGMD \cite{Wang2018}, along with recent work on the substantial segregation of ON-excitation to field C of the LGMD1 \cite{Dewell2022} all informed the significance of polarity motion vision in locust's visual systems. 
	Accordingly, considering the homology between different insects' visual brains, and the universal findings of polarity vision across researched animals' dynamic vision systems, the simulation of LGMD2's circuits, and separation of ON and OFF based FFI provide strong hypothesis to neuroscience.
    
    In recent studies, the competition between ON/OFF channels also indicate how opposite-polarity signals could interact to suppress translating-induced excitation in order to sharpen up the looming selectivity of LGMD \cite{Lei2023}. 
    Furthermore, the performance of LGMD-based real-world applications is partially restricted in noisy and low-contrast environments. 
    Keil's computational model suggested that threshold-involved neuronal signal integration could pool signals from different channels, thereby enhancing looming selectivity in noisy environments \cite{Keil2015}. 
    By analyzing realistic synaptic integration within bio-physically accurate neuronal models, Poleg-Polsky found that introducing an additional threshold for dendritic spikes significantly increased neuronal tolerance to a wide variety of noise \cite{Alon2019}. 
    Besides, the recent studies proposed that normalizing photoreceptor signals can enhance LGMD responses to low-contrast stimuli \cite{Fu2023b, Fu2023c}. 
    Similarly, global inhibition observed in the locust's visual pathways helps maintain excitation within the dynamic range of the neuron, thereby preserving functional selectivity under varying illumination conditions \cite{Zhu2018}. 
    These mechanisms are also present in other sensory circuits, highlighting their fundamental role in neural processing \cite{Carandini2012}. 
    Moreover, the deepened LGMD1-LGMD2 cascade network shows selectivity to only approaching targets which might reveal some cooperative mechanisms interacting the pre-synaptic dendrites between LGMD1 and LGMD2, nevertheless unknown in terms of physiology \cite{Fu2023b,Rind2014a}.

    Unlike the fruit fly \textit{Drosophila}, the biological mechanisms underlying the visual pathways responsible for looming perception in locusts remain significantly under-explored, leaving many unanswered questions and arguments, as highlighted in the previous section. 
    In this context, bio-inspired modeling and robotic approaches offer a valuable means to leverage existing knowledge about LGMDs, providing practical frameworks to simulate, validate, and extend our understanding of these neuronal processes.
	
\section{Discussion}
	Physiological and anatomical discoveries of the LGMD neurons have undoubtedly inspired the development of parsimonious and efficient collision detection methods, many of which have been successfully applied to mobile robots, both on the ground and in the air. 
    The behavior of these LGMD-based robots has validated the established neuroscience and anatomical findings, while the limitations observed in complex real-world scenarios may suggest the presence of yet-undiscovered mechanisms or biological substrates within the LGMD neurons or its afferent neural circuits. 
    The bio-inspired research paradigm of LGMDs has demonstrated considerable success in integrating neuroscience and robotics. 
    We propose that this success can be attributed to the following two factors.
	
	\textit{Application demands} - First, there is an urgent need for efficient, low-energy collision detection methods in real-world applications. 
    Collisions are ubiquitous and pose significant hazards to both living creatures and intelligent robotic systems. 
    Current collision detection methods are often energy-intensive due to the need for map reconstruction and object recognition \cite{Perez2017, Cigla2017, Garcia2020}, or they rely heavily on specific sensors such as radar or wireless ultraviolet light \cite{Reich2020, Zhao2022}. 
    Insects, with their tiny brains and highly effective collision avoidance behaviors, offer an ideal model for developing parsimonious and efficient collision detection systems. 
    The evolved, specialized structure of looming-selective neural circuits in insects provides a foundation for designing straightforward models capable of addressing real-world collision challenges. 
    There are two identified types of neurons or neuron classes that respond selectively to approaching objects: the LGMDs (one LGMD1/LGMD2 per eye in grasshoppers) and the lobula plate/lobula columnar type-2 (LPLC2) neurons ($\sim 80$ LPLC2 neurons per eye in flies). 
    LPLC2 neurons are located in the lobula complex of \textit{Drosophila}, and unlike LGMD, their excitation relies on the precise spatial arrangement of dendrites within a directionally selective motion feature map provided by T4/T5 neurons \cite{Klapoetke2017, Ache2019}. 
    LPLC2 neurons prefer motion that is outward-from versus inward-toward their receptive field center, and this radial motion opponency causes LPLC2 to be excited by looming objects at the center of the receptive field rather than by receding objects. 
    However, this characteristic limits LPLC2's response in near-miss scenarios or to looming patterns from different directions, making it potentially insufficient for handling complex real-world challenges. 
    In contrast, LGMD neurons can detect imminent collisions from various directions in both stationary and moving backgrounds, without requiring directionally specific afferent signals \cite{Jones-Gabbiani-2010,Yakubowski2016}. 
    The robustness of LGMD responses makes it a suitable alternative for generating accurate yet efficient collision detection methods.
	
	\textit{Biological prominence} - Second, comprehensive research in neuroscience and anatomy on the LGMD neuron has significantly contributed to the successful development of artificial systems. 
    The LGMD neuron is large in size, with an extensive dendritic structure that makes it easier to study using various experimental protocols. 
    Dendritic field A spans approximately $7705 \mu m$ in length and $108,400 \mu m^2$ in surface area, while dendritic fields B and C are shorter, with lengths of approximately $4438 \mu m$ and $2517 \mu m$, surface areas of $28,100 \mu m^2$ and $13,500 \mu m^2$, and radii of $0.8 \mu m$ and $0.7 \mu m$, respectively \cite{Gabbiani2007}. 
    These lengths and areas are for the combined length of all dendritic branches. 
    This morphology facilitates research using tools such as extracellular hook electrodes, two-photon microscopy, and optogenetics \cite{Gabbiani2005b, Gabbiani2007, Zhu2016, Wang2018b}. 
    Furthermore, investigating the neuronal response of the LGMD is particularly feasible in laboratory settings because approaching objects can be effectively simulated using two-dimensional projections. 
    These factors have sustained the vibrant wave of neuroscience and anatomical research on LGMD neurons since their discovery in the 1970s. 
    Over time, this research has become increasingly comprehensive, addressing aspects ranging from morphology to membrane conductance \cite{Dewell2018b, Dewell2019}. 
    Consequently, it has provided an abundant source of inspiration for developing LGMD-based applications in real-world scenarios.
	
	Similar motion perception neurons found in insect visual systems, such as lobula plate tangential cells (LPTCs) and small-target motion detectors (STMDs), may also show potential on generating this paradigm. 
    LPTCs, located in the lobula plate of the fly neuropile, function as wide-field detectors that integrate upstream visual signals from T4 and T5 cells in the medulla and lobula, as well as L1 (ON) and L2 (OFF) interneurons in the lamina. 
    These cells have extensive dendritic arborizations, allowing them to process motion signals across a broad visual field and respond selectively to horizontal or vertical movement \cite{Schnell2012, Cole2013}. 
    The mechanisms underlying LPTCs have inspired motion detection algorithms in robotics, particularly for applications in optic flow-based navigation and obstacle avoidance \cite{Serres2017, Li2021}. 
    STMDs are characterized by their sensitivity to small-target motion, with peak responses to targets subtending $1 \sim 3 \degree$ of the visual field. 
    STMD neurons have been observed in several insects, including hoverflies and dragonflies \cite{Collett1975, Carroll1993, Wiederman2008}. 
    Over the past decades, STMD-based principles have demonstrated potential applications in hardware and UAV technologies \cite{Halupka2011, Bagheri2017, Wang2023}.
	
	Although the neuronal response of LGMDs can be effectively simulated using the NEURON simulation environment\footnote{https://www.neuron.yale.edu/neuron/} \cite{Dewell2018}, the afferent mechanisms driving LGMD responses to different types of motion remain largely unexplored. 
    Olson et al. \cite{Olson2021} focused on replicating the neural signal processing of LGMD by incorporating global, lateral, and feedforward inhibition, demonstrating alignment between model outputs and real LMC responses. 
    However, their model was limited in scope, considering only the OFF pathway while excluding the ON pathway, ON/OFF interactions, as well as self-inhibition and lateral excitation. 
    The limited scope covered by Olson et al. is a sign that there is more work to be done towards full processing dynamics of LGMDs.
    
    Moreover, the connectivity between LGMD neurons and other neurons within the locust neuropile remains largely undefined. Recent connectomic data from \textit{Drosophila} suggested that integrating whole-field looming detection pathways with local directional signals can significantly enhance collision detection accuracy in both simulated and real-world environments \cite{Vashistha2022, Song2024}. 
    Additionally, modelers have combined LGMD1 with LPLC2 to improve looming perception performance in real-world scenarios, showing significant promise for high-speed collision detection tasks \cite{Fu-ARM2020,Zhao2023b}. 
    Another unresolved question concerns the evolutionary rationale behind the emergence of distinct LGMD1 and LGMD2 neurons, each with unique morphology. 
    Understanding why and how these specialized neurons developed over thousands of years is critical for replicating their adaptability and efficiency in engineered models.
	
	These unresolved issues present substantial challenges to the development of biologically plausible LGMD models and hinder the establishment of effective LGMD-based collision detection systems. 
    While LGMD-based models often achieve timely and accurate collision detection through supplementary mechanisms—such as gradient-based spike adaptation frequency or average-value denoising layers—they still fall short of replicating the adaptability and robustness exhibited by real LGMD neurons. 
    In nature, collision avoidance is achieved through inherent biological structures and rapid signal transmission without the need for complex computational processing. 
    Current LGMD-based models thus face a trade-off between replicating comprehensive, biologically accurate neural structures and designing efficient, real-world-ready signal processing systems. 
    While these models have succeeded in validating some neuroscience findings, they ultimately fall short in providing the adaptability and robustness of biological systems, often leaving neuro-scientists with more questions than answers—highlighting areas for improvement rather than offering mature, practical solutions.

    Fortunately, recent advancements in neuroscience research on \textit{Drosophlia} have reached a groundbreaking milestone with the complete reconstruction of the entire \textit{Drosophlia} brain connectome \cite{Schlegel2024,Dorkenwald2024,Matsliah2024,Shiu2024}. This significant achievement is poised to illuminate neuroscience research across all insect species. Notably, the dense reconstruction of all neurons in the anterior visual pathway (AVP) provides critical insights into how \textit{Drosophlia} encodes and integrates visual information \cite{Garner2024}. Given the anatomical similarities between the optic lobes of locusts and \textit{Drosophlia}, including the presence of homologous neurons, this reconstruction offers a valuable foundation for understanding the complete neural signaling pathway to the LGMDs. Furthermore, it deepens our understanding of LGMDs at the levels of individual neurons and their connectome.
    With these insights and inspirations, LGMD-based computational modeling and robotics will inevitably enter a new era, offering more biologically plausible and parsimonious solutions for collision detection.

	While we are still far from fully understanding the intrinsic mechanisms that enable the locust nervous system to achieve its remarkable collision detection capabilities, even a simple replication of the LGMD neuronal response characteristics has already proven sufficient for developing low-energy, robust collision detection methods. 
    This bio-inspired paradigm exemplifies a fundamentally different methodology compared to deep learning-based collision detection strategies. 
    Deep learning methods, though originally inspired by brain functions, primarily emphasize the learning process. 
    They derive their strength from uncovering intrinsic patterns and features within extensive training datasets, achieving outstanding performance through experience-based adaptation. 
    In contrast, LGMD-based collision detection methods focus on directly replicating neuronal responses and neural architectures, providing an alternative approach for real-world collision detection that operates without relying on learning or large datasets. 
    This bio-inspired paradigm of LGMD offers a promising pathway for efficient collision detection, especially in scenarios where simplicity, low computational overhead, and rapid responses are crucial.
    We believe these two paradigms—deep learning-based and LGMD-based approaches—can complement and enhance each other. Deep learning methods can provide information about approaching objects and the agent's self-location, further enhancing the performance of LGMD-based collision detection methods, particularly in complex real-world scenarios. Conversely, LGMD-based methods can act as supplementary mechanisms to support deep learning models in situations where training data are limited or unavailable. Together, these approaches form a synergistic framework for advancing collision detection technologies.
	
	
 \section{Conclusion}
	To sum up, this review illustrates an LGMD-based research paradigm that bridges neuroscience, computational modeling, and bio-inspired robotics. 
    Physiological and anatomical research has inspired computational models for real-world applications, subsequently enhancing the capability of mobile robots to interact effectively, safely with their physical environments. 
    The behaviors of these robots in real-world scenarios not only validate the mechanisms and substrates identified in neuroscience and anatomy but also suggest the potential for undiscovered connections or functional substrates. 
    The advancement of this research paradigm is driven by a strong demand for practical applications, and supported by solid foundation of neuroscience and anatomical studies. 
    Similarly, other single neurons or neuronal systems with comparable conditions hold the potential to establish a similar, cyclical, and mutually reinforcing research framework.
	
	\backmatter

	%
	%

	\noindent
	

	
	\bibliography{main.bib}

\end{document}